\documentclass[conference]{IEEEtran}
\IEEEoverridecommandlockouts

\usepackage{cite}
\usepackage{graphicx}
\usepackage{psfrag} 
\usepackage{bm}
\usepackage{subfigure}
\usepackage{amsmath,amssymb,amsfonts}
\usepackage{epsfig}
\usepackage{amsmath}
\usepackage{multirow}
\usepackage{lineno}
\usepackage{algorithm}
\usepackage{algorithmic} 
\usepackage{diagbox} 
\usepackage[margin=0.7in]{geometry}
\usepackage{lipsum}

\usepackage{xcolor}

\newcommand{\p}[1]{\mathop{\mbox{\it p} } }

\renewcommand{\vec}[1]{\ensuremath{\boldsymbol{#1}}}
\newcommand{\be}{\begin{equation}}
\newcommand{\ee}{\end{equation}}
\newcommand{\ba}{\begin{array}}
\newcommand{\ea}{\end{array}}
\newcommand{\bea}{\begin{eqnarray}}
\newcommand{\eea}{\end{eqnarray}}
\newcommand{\bean}{\begin{eqnarray*}}
\newcommand{\eean}{\end{eqnarray*}}

\newcommand{\rmt}{^{\rm T}}

\definecolor{white}{rgb}{1,1,1}

\newtheorem{theorem}{Theorem}

\newtheorem{property}{Property}
\newtheorem{definition}{Definition}

\title{Invariant Transformation and Resampling based Epistemic-Uncertainty Reduction}

\author{Sha Hu\\
Lund Research Center, Huawei Technologies Sweden AB, Sweden.\\ 
Email: hu.sha@huawei.com
}

\begin{document}
\maketitle

 \vspace{-2mm}

\begin{abstract}
An artificial intelligence (AI) model can be viewed as a function that maps inputs to outputs in high-dimensional spaces. Once designed and well trained, the AI model is applied for inference. However, even optimized AI models can produce inference errors due to aleatoric and epistemic uncertainties. Interestingly, we observed that when inferring multiple samples based on invariant transformations of an input, inference errors can show partial independences due to epistemic uncertainty. Leveraging this insight, we propose a ``resampling“ based inferencing that applies to a trained AI model with multiple transformed versions of an input, and aggregates inference outputs  to a more accurate result. This approach has the potential to improve inference accuracy and offers a strategy for balancing model size and performance.
\end{abstract}

 \vspace{-2mm}
\section{Introduction}

Artificial intelligence (AI) and machine learning (ML) are emerging as key enablers~\cite{QW24, Mehran2019Deep, Jiang2021Dual, Sun2020Learn, SC23, Raviv2023Modular, Zhou2021RCNet} in sixth-generation (6G) system. Recent proposals aim to enhance conventional signal processing modules, inducing channel estimation (CE)~\cite{MK21, LL17, SS19, LT23} and multi-input multi-output (MIMO) detection~\cite{Schmid2022LowComplexity, Honkala2021DeepRx, Faycal2022E2E}. While supervised learning allows AI models to be trained with labels, there are two uncertainties in ML, known as aleatoric uncertainty and epistemic uncertainty~\cite{HZ23, JW25}, respectively. Aleatoric uncertainty reflects the inherent randomness from the noisy data and is reducible, while epistemic uncertainty measures the imperfectness of learning and can be reduced by means of scaling training dataset and model size, optimizing model architecture, and adjusting learning methods. Due to uncertainties, even a well-trained AI model with large network size can still produce inference errors.

In the context MIMO detection, conventional quasi maximum-likelihood (ML) detectors~\cite{Zhang2022SoftMIMO, Wubben2004MMSE, Im2007QRMMLD, SF2017} perform near-optimally. AI based designs have also demonstrated comparable performance~\cite{QW24}. However, as system grows more complex when signal-to-noise ratio (SNR) increases, such as with large MIMO sizes (e.g., $8\!\times\!8$) and high-order quadrature amplitude modulation (QAM) like 64QAM and 256QAM, suboptimal detectors can suffer from performance losses relative to the ML detector. These losses become pronounced with a high code-rate when operational uncoded bit error rate (BER) is below 1\%. This presents a challenge for AI based MIMO detector that relies on learning from statistical patterns in training data. Achieving such a high precision requires very large networks and extensive amount of training data, leading to increased complexity and cost.

Instead of optimizing AI model in training process, we consider an interesting question: {\it Can inference accuracy be improved without a trained AI model?} Our findings suggest the answer is yes. The idea is to leverage properties of the considered problem that are difficult for AI to fully capture from learning, such as invariant transformations of MIMO detection model. Specifically, when performing inferences with multiple samples, inference errors exhibit partial statistical independence that arises from the epistemic uncertainty. This can be exploited to enhance a trained AI model  through a ``resampling'', where multiple inferences are treated as different samples for the trained AI model. A combination of resampled inference outputs can yield a better results with less epistemic uncertainty. Fig.~1 illustrates this concept, and we provide analytical insights and numerical results to demonstrate the enhancements with resampling for AI based MIMO detection.

Note that a similar framework has been known as test time augmentation (TTA)~\cite{SG21, CP22, KB24} in image classifications, which makes predictions on a set of inputs at test time with data augmentations and aggregates those predictions for a final prediction. However, data augmentations employed in TTA predominantly rely on heuristic techniques such as image rotations, flips, and adding extra noise, which lack solid theoretical justifications. Furthermore, the optimal strategy for combining multiple outputs remains ambiguous. By comparison, our ``resampling'' proposal is grounded in invariant transformations and endowed with clear mathematical meanings, and the optimal combining can be derived from error correlations. Moreover, we also elucidate the connections between resampling a trained AI model and epistemic uncertainty reduction.

\begin{figure}
    \centering
                \hspace{2mm}
    \includegraphics[width = 0.4\textwidth, keepaspectratio]{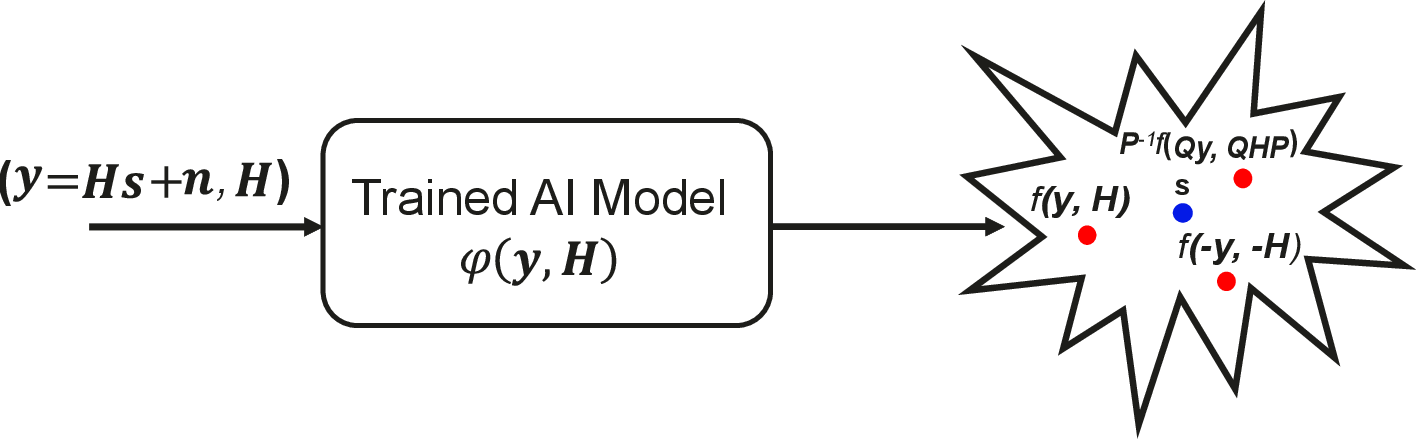}
        \vspace{-4mm}
    \caption{Resampling a trained MIMO detector with multiple inference samples generated from conjugates, flips, and permutations for an input $(\vec{y}, \vec{H})$.}  
    \label{fig:MIMO}
    \vspace{-7mm}
\end{figure}

 \begin{figure*}
    \centering
                \hspace{2mm}
    \includegraphics[width = 0.65\textwidth, keepaspectratio]{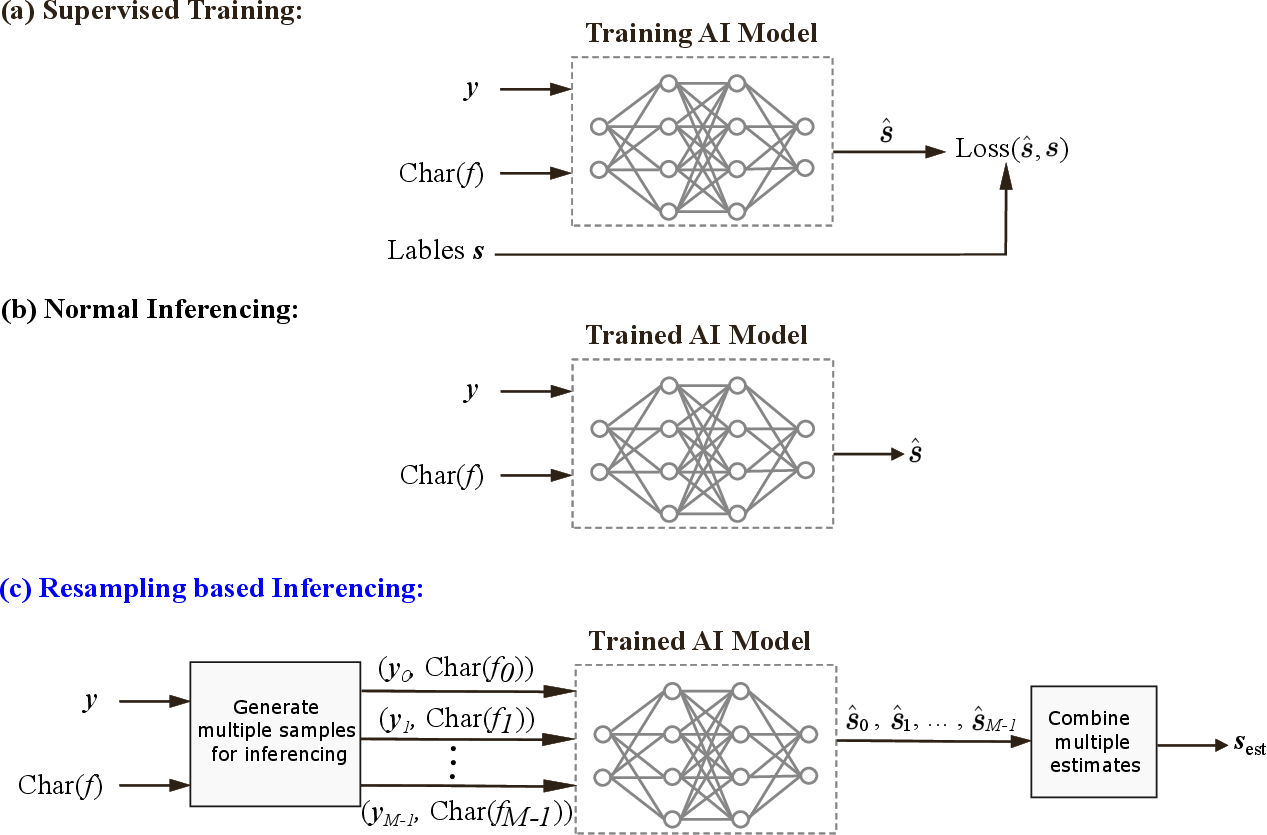}
        \vspace{-2mm}
    \caption{Conventional supervised-learning based training (a) and inferencing (b), and the proposed resampling based inferencing with multiple samples generated from an input observation. In case $f$ is unknown,  $\text{Char}(f)$ is excluded from the inputs to the model.}  
    \label{fig:AIRec}
    \vspace{-6mm}
\end{figure*} 

 \vspace{-2mm}
\section{Resampling A Trained AI model}  

\subsection{Mathematical Formulations}  

Consider a general input-output system
\bea \label{md1} \vec{y} = f(\vec{s})+\vec{n},\eea
where the hyper-dimensional parameter $\vec{s}$ is the input and $\vec{y}$ is the output corrupted by a noise $\vec{n}$. Without loss of generality, the noise is assumed zero-mean and the variance is normalized to one and excluded from the model inputs. The system is governed by a mapping function $f$, which can be linear or non-linear, known or unknown before hand.

A common task is to estimate $\vec{s}$ from a noisy observation $\vec{y}$ as in (\ref{md1}). In this case, the optimal Bayesian estimator that maximizes a posteriori  (MAP) is
\bea \label{map} \hat{\vec{s}} = \text{argmax} \; p(\vec{y}|\vec{s})p(\vec{s}), \eea
where $p(\vec{y}|\vec{s})$ is the likelihood and $p(\vec{s})$ is the priors of $\vec{s}$. In the absence of $p(\vec{s})$, it reduces to the ML estimator
 \bea \label{ml} \hat{\vec{s}} = \text{argmax} \; p(\vec{y}|\vec{s}). \eea

In general, the noise $\vec{n}$ is also modeled as additive white Gaussian noise (AWGN). Under this assumption, maximizing the likelihood $p(\vec{y}|\vec{s})$ is equivalent to minimize the error
 \bea \label{mse} \hat{\vec{s}} = \text{argmin} \; \|\vec{y} - \vec{f}(\vec{s}) \|^2 . \eea
In high dimensional spaces, an exhaustive search over $\vec{s}$ is computationally infeasible. For instance, in an $8\!\times\!8$ MIMO system with 256QAM modulation, the search size for $\vec{s}$ is $8^{256}$, which is prohibitively large. Suboptimal algorithms have been developed over the past half century under the umbrella of MIMO detection, offering near-optimal performance with significantly reduced complexity.

With the rise of AI, it is also possible to apply supervised-learning to solve (\ref{md1}). Essentially, the AI model is trained to learn an inverse mapping
\bea \label{invmap} \hat{\vec{s}} = \varphi(\vec{y}, \text{Char}(f)), \eea
where $\text{Char}(f)$ represents the mapping characteristic of $f$.
\begin{definition}[Mapping Characteristic]
The characteristic $\text{Char}(f)$ is defined as a representation of mapping $f$ that serves as an input to the AI model associated with the observation $\vec{y}$ in training and inferencing for solving (\ref{md1}). 
\end{definition}

The form of $\text{Char}(f)$ is design-dependent. For example, in a linear MIMO system 
\bea \label{mm} \vec{y} = \vec{H}\vec{s}+\vec{n} \eea
where $f(\vec{s})=\vec{H}\vec{s}$, the characteristic can be defined as $\text{Char}(f)\!=\!\vec{H}$ or $(\text{Re}\{\vec{H}\}, \text{Im}\{\vec{H}\})$ if $\vec{H}$ is complex-valued. 

An AI based MIMO detector operates in two stages: a training stage and an inference stage, as illustrated in Fig.~2(a) and Fig.~2(b), respectively. During training stage, a sufficiently large dataset comprises of input-output pairs $(\vec{y},\; $\vec{s}$)$ is used to train the model, together with the characteristic $\text{Char}(f)$. The inputs $\vec{s}$ serve as labels to minimize a predefined loss function. After the model is well-trained, it is applied in inference, where it estimates $\hat{\vec{s}}$ for any observation $\vec{y}$, most of which are previously unseen in training. The AI model is expected to exploit the statistics from the training data and learn a reverse-mapping $\varphi(\cdot)$ that reconstructs $\vec{s}$ from $\vec{y}$ and $\text{Char}(f)$.

\subsection{Invariant Transformations}  

Although AI based designs can be effective, one limitation is that they often fail to fully learn the inherent properties of the system. Consider a general transformation $\mathcal{T}(\vec{y})$ on (\ref{md1}) with two functions $g$ and $q$ (and an inverse function $q^{-1}$) to form a new mapping $ \mathcal{T}_f\!=\!g\circ f\circ q^{-1}$ such that
\be \label{md2} \mathcal{T}(\vec{y}) = (g\circ f\circ q^{-1})(q(\vec{s}))+g(\vec{n})= \mathcal{T}_f(q(\vec{s}))+g(\vec{n}).\ee

\begin{definition}[Invariant Transformation]
For a transformation $\mathcal{T}$ in (\ref{md2}) to be invariant, it shall meet the condition that the distributions of $\text{Char}(\mathcal{T}_f)$, $q(\vec{s})$, and $g(\vec{n})$ are all identical to $\text{Char}(f)$, $\vec{s}$, and $\vec{n}$, respectively.
\end{definition}

For instance, if the channel $\vec{H}$ in (\ref{mm}) follows a Rayleigh distribution~\cite{QW24}, and the entries in $\vec{s}$ drawn from a QAM constellation are independent and identically distributed (i.i.d.), then the following transformations are invariant:
\begin{itemize}
 \item Linear transformation with a unitary matrix $\vec{Q}$:
\bea \label{mm_it1} \mathcal{T}(\vec{y})=\vec{Q}\vec{y} =\vec{Q} \vec{H}\vec{s}+\vec{Q}\vec{n}. \eea
 \item Permuting channel and data entries via a permutation matrix $\vec{P}$:
\bea \label{mm_it1}  \mathcal{T}(\vec{y})=\vec{y} =\big(\vec{H}\vec{P}\rmt)\big (\vec{P}\vec{s})+\vec{n}. \eea
 \item Complex conjugate:
\bea \label{mm_it2}  \mathcal{T}(\vec{y})=\vec{y}^{\ast} = \vec{H}^{\ast} \vec{s}^{\ast}+\vec{n}^{\ast}. \eea
\end{itemize}
Specifically, invariant transformations including unitary rotations, conjugate, and permutations arise from the statistical properties of $\vec{H}$, $\vec{s}$, and $\vec{n}$. These operations preserve the statistical structure of system, and lead to equivalent detection performance in ML sense. Hence, the below Property~1 holds.

\begin{property}
For an AI model trained on input samples \big($\vec{y}$, $\text{Char}(f)$\big) and labels $\vec{s}$, it also works for transformed samples $\big(\mathcal{T}(\vec{y}), \text{Char}(\mathcal{T}_f)\big)$ under an invariant transformations $\mathcal{T}$, and with equal inference accuracy.
\end{property}

\subsection{Resampling Theorem}

Next we present the main result.  In general, an inference output from the trained AI model can be modeled as
\bea \label{set} \hat{\vec{s}} = \vec{s}+\vec{z}, \eea
where $\vec{z}$ represents an estimation error and can be reasonably modeled as Gaussian due to the Central Limit Theorem. If there are multiple estimates of $\vec{s}$, the accuracy can be refined.

\begin{theorem}
Assume there are $M$ estimates of a scalar variable $s$, denoted as $s_m \!=\! s \!+\! z_m$, and $z_m$ represents a zero-mean estimation error with identical variance $\text{var}(z_m)\! =\! \sigma^2$. Further, the correlation between any two errors satisfies $\text{cov}(\vec{z}_m,\vec{z}_n) \!=\! \rho_{mn} \sigma^2$ for $m\ne n$. Construct an estimate $\bar{s}$ with a linear combination
\bea \label{sbar}
\bar{s} = \sum_{m=0}^{M-1} \beta_m s_m \eea
subject to $\sum_{m=0}^{M-1} \beta_m \!=\! 1$. The minimum variance of estimation error in $\bar{s}$ is
\bea  \text{var}(\bar{s}) =  \frac{1}{\vec{1}\rmt\vec{R}^{-1}\vec{1}}, \eea
 attained with weights 
 \bea \vec{\beta}\!= \frac{\vec{R}^{-1}\vec{1}}{\vec{1}\rmt\vec{R}^{-1}\vec{1}}. \eea
Here $\vec{R}$ denotes the covariance matrix among the $M$ estimates and \vec{1} is the vector with all entries equal to 1. The vector $\vec{\beta}$ comprises of combination weights in (\ref{sbar}).
\end{theorem}

Theorem~1 can be directly verified. As an example, considering the case $\rho_{mn}\!=\!\rho$ for $m\!\ne\!n$, the optimal weight $\beta_m\!=\!1/M$ and the minimum variance is $\rho\!+\!\frac{1-\rho}{M}$. 

With Property~1, $M$ inference samples can be generated from transformations with an input $(\vec{y}, f)$. This can be viewed as ``resampling'' of the function $\varphi(\vec{y}, f)$, as shown in Fig.~2(c), and it turns out to be an effective technique when $\rho\!<\!1$. Note that as $M\!\to\!\infty$, the minimum variance of estimation error converges to $\rho$. An illustration of variance decrement with refined estimate is shown in Fig.~3.

\subsection{Epistemic Uncertainty Reduction with Resampling}  

The inference error in (\ref{set}) contains two parts: aleatoric uncertainty inherited from the randomness in $p(\vec{s}|\vec{y})$, and epistemic uncertainty due to imperfect learning. From~\cite{JW25}, the epistemic uncertainty can be measured by
\bea   \label{epi} \text{var}_{\mathcal{T}}\big(\mathbb{E}_{\vec{s}|\vec{y}}[\hat{\vec{s}}|\mathcal{T}(\vec{y}),\mathcal{T}_f]\big) \!=\! \mathbb{E}_{\mathcal{T}}\big[\big(\varphi_{\mathcal{T}}(\mathcal{T}(\vec{y}), \mathcal{T}_f)-\bar{\varphi}_\mathcal{T}\big)^2\big]\!. \eea
For simplicity, here we assume that for any transformation $\mathcal{T}(\vec{y})$, the inference result is $\vec{s}$. The mapping function $\varphi_{\mathcal{T}}$ denotes a trained network with the training dataset transformed with $\mathcal{T}$. Its inference output is denoted as $\varphi_{\mathcal{T}}(\mathcal{T}(\vec{y}), \mathcal{T}_f)$ for the observation $\mathcal{T}(\vec{y})$. The mean of inference results from all transformations that removes epistemic uncertainty equals
\bea \bar{\varphi}_\mathcal{T} =\mathbb{E}_{\mathcal{T}}[\varphi_{\mathcal{T}}(\mathcal{T}(\vec{y}), \mathcal{T}_f)].\eea 

On the other hand, with the proposed resampling technique, in essence it constructs a new mapping function
\bea \label{bphi} \underline{\varphi}_\mathcal{T} =\mathbb{E}_{\mathcal{T}}[\varphi(\mathcal{T}(\vec{y}), \mathcal{T}_f)].  \eea 

\begin{property}
If the training dataset \{$\vec{y}$\} contains all invariant transformations such that $ \{\mathcal{T}{\vec{y}}\}\!=\! \{\vec{y}\}$ for any $\mathcal{T}$, then it holds that $\varphi_{\mathcal{T}}\!=\!\varphi$, and
\bea 
 \bar{\varphi}_\mathcal{T}=  \underline{\varphi}_\mathcal{T}.  \eea 
\end{property}

Property~2 reveals that a training dataset needs to be infinitely large to mitigate epistemic uncertainty. However, (\ref{bphi}) provides another approach to mitigate epistemic uncertainty in inference stage other than in training~\cite{HZ23, JW25}, especially when the difference between $\underline{\varphi}_\mathcal{T}$ and $\bar{\varphi}_\mathcal{T}$ is small.  On the other hand, by increasing the number of samples $M$, the ``resampling based inference'' can serve as a performance bound for the AI model to be trained, in the sense of minimizing epistemic uncertainty.

\begin{figure}[t]
    \centering
            \hspace{4mm}
    \includegraphics[width = 0.31 \textwidth, keepaspectratio]{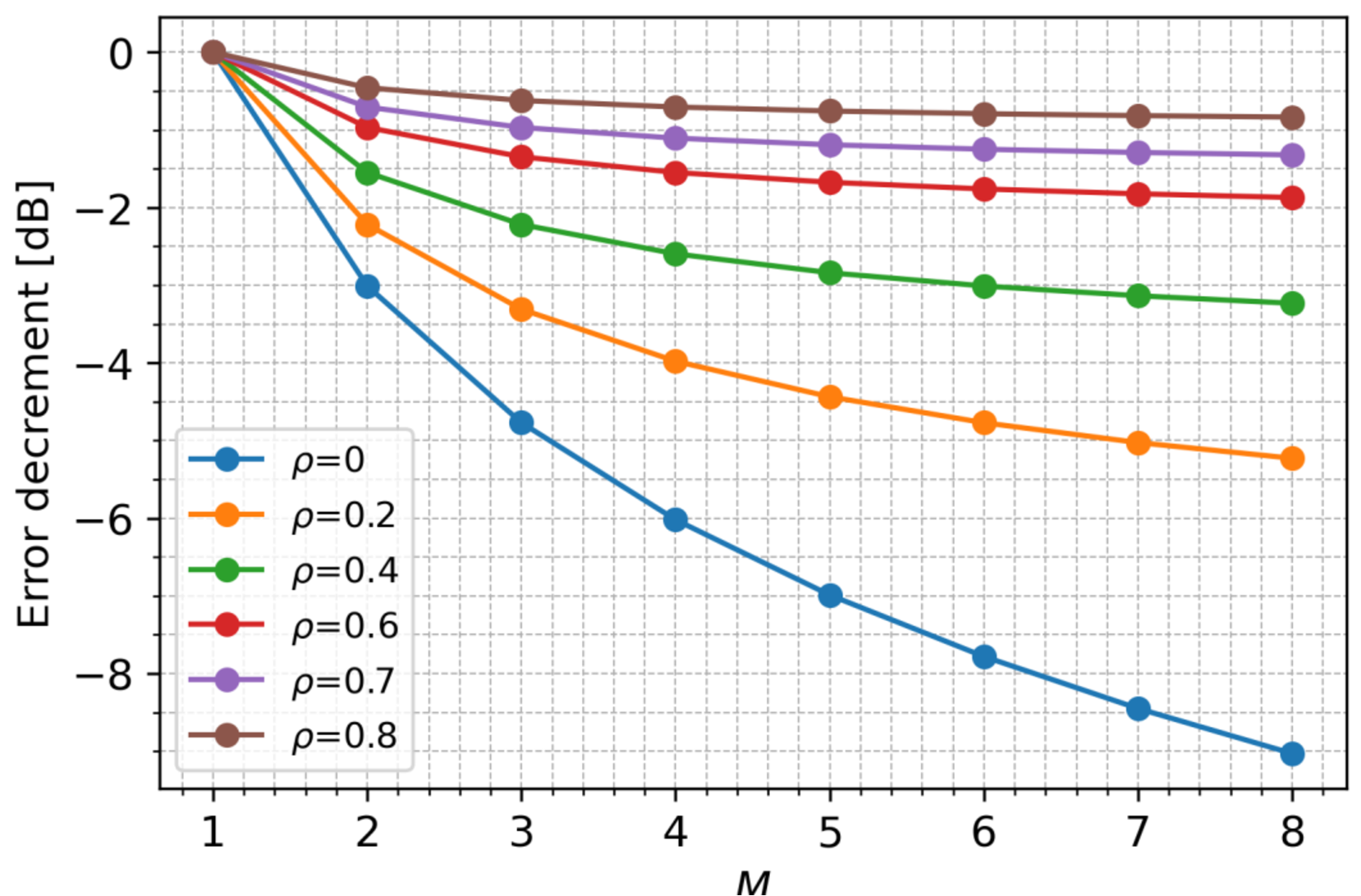}
        \vspace{-3.5mm}
    \caption{The variance decrements in relation to $\rho_{mn}\!=\!\rho$ for $m\!\ne\!n$.} 
    \label{fig:AIRec}
    \vspace{-7mm}
\end{figure} 

\subsection{Discussions}  

When noise level in the system is high in (\ref{md1}), aleatoric uncertainty dominates the inference error in (\ref{set}), and this often results in high correlations across inference errors from different transformations, thereby limiting the benefits of resampling. In contrast, as noise level goes down, epistemic uncertainty becomes dominant in the inference error, making resampling more effective at high SNR.

On the other hand, in case $f$ or its properties are unknown, since generating multiple samples via invariant-transformation is not straightforward, heuristic resampling methods can be considered. One heuristic resampling can construct transformed or perturbed versions based on $\vec{y}$ and inference these samples with the same AI model. The inference results can be combined either via a conventional method or with another neural network. The heuristic method sometimes can be effective, but unlike resampling with invariant transformations, an analytical understanding can be intractable.


\section{Simulation Results with MIMO Detection}

Consider a MIMO transmission on two resource blocks in 5G system. An AI based MIMO detector is constructed with Transformer Encoder with inputs $(\vec{y}, \vec{H})$. The output is a matrix comprises of either log-likelihood ratio (LLR) values $L_{m,k}$ for the $k$th bit on the $n$th layer, or the marginal probabilities $p(s_{n,m} \!=\!s_ m|\vec{y}, \vec{H})$ for each constellation symbol $s_m$ on each layer. The AI model contains four Transformer Encoders and each with 4 heads. The model size and feed-forward network (FFN) sizes are set to 256 and 1024, respectively, for $4\!\times\!4$ MIMO. This yields a number of trainable parameters close to $3M$, and both sizes are doubled for larger MIMO and higher-order modulations.

\begin{figure}
          \hspace*{-2mm}
    \centering
      \hspace{-4mm}
    \includegraphics[width = 0.35 \textwidth, keepaspectratio]{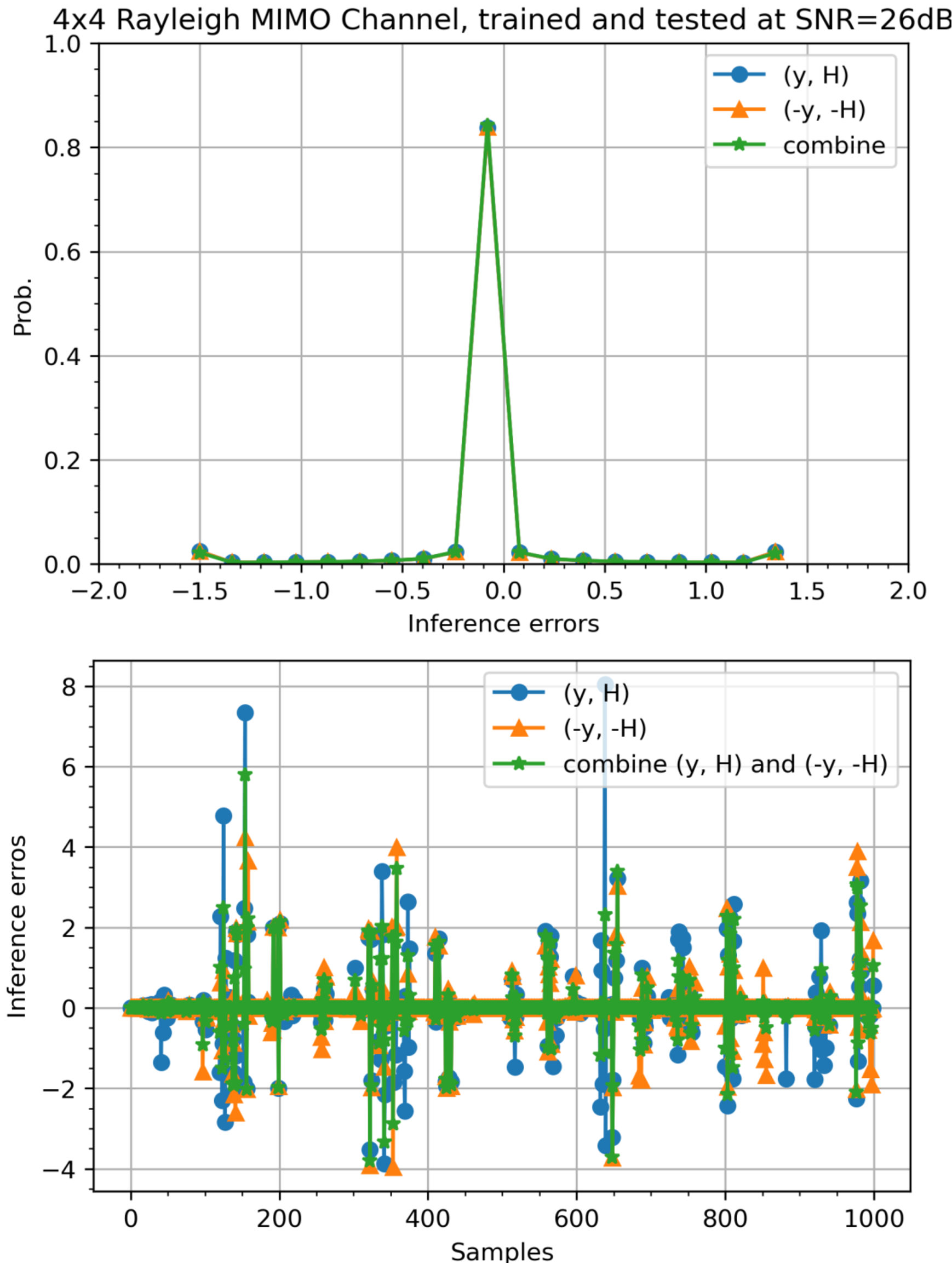}
        \vspace{-3mm}
    \caption{The distribution of inference errors and example samples. }  
    \label{fig:AIRec}
    \vspace{-6mm}
\end{figure}

\subsection{Inference Errors}
We firstly evaluate a $4\!\times\! 4$ MIMO with 64QAM modulation in (\ref{mm}), and $\vec{H}$ is modeled as i.i.d. Rayleigh channel. Since the AI model applies real-valued inputs, the complex-valued model is equivalently converted into an $8\!\times\! 8$ real-valued model, and $\vec{s}$ contains 8PAM symbols uniformly drawn from $\{\pm1, \pm3, \pm5, \pm7\}$. The AI detector is trained at SNR=26dB and outputs the marginal probabilities $p(s_{n,m} \!=\!s_ m|\vec{y}, \vec{H})$. The measured SERs when inferencing with $(\vec{y}, \vec{H})$ or $(-\vec{y}, -\vec{H})$ are both equal to 5.7\%, which matches with SER in training. Notably, if these two inference outputs are combined, SER is reduced to  5.1\%.

Fig.~4 illustrates the inference errors $z$ in reconstructed soft estimates $\mathbb{E}(s_ m|\vec{y}, \vec{H})$,  which is truncated into interval (-1.5, 1,5), since an symbol error occurs when an error fall outside this range. Inference errors corresponding to (\vec{y}, \vec{H}) and (-\vec{y}, -\vec{H}) both approximately exhibit Gaussian distributions, with a mean-value 0.004 that is close to zero and a variance $\sigma^2\!=\!0.385$. The correlation between these two errors is $\rho\!=\! 0.71$. After averaging the two inference outputs, the error-variance is reduced to 0.33, which perfectly aligns with the prediction $(\rho\!+\!\frac{1-\rho}{M})\sigma^2$ based on Theorem~1.

\begin{figure}
    \centering
          \hspace*{8mm}
    \includegraphics[width = 0.38 \textwidth, keepaspectratio]{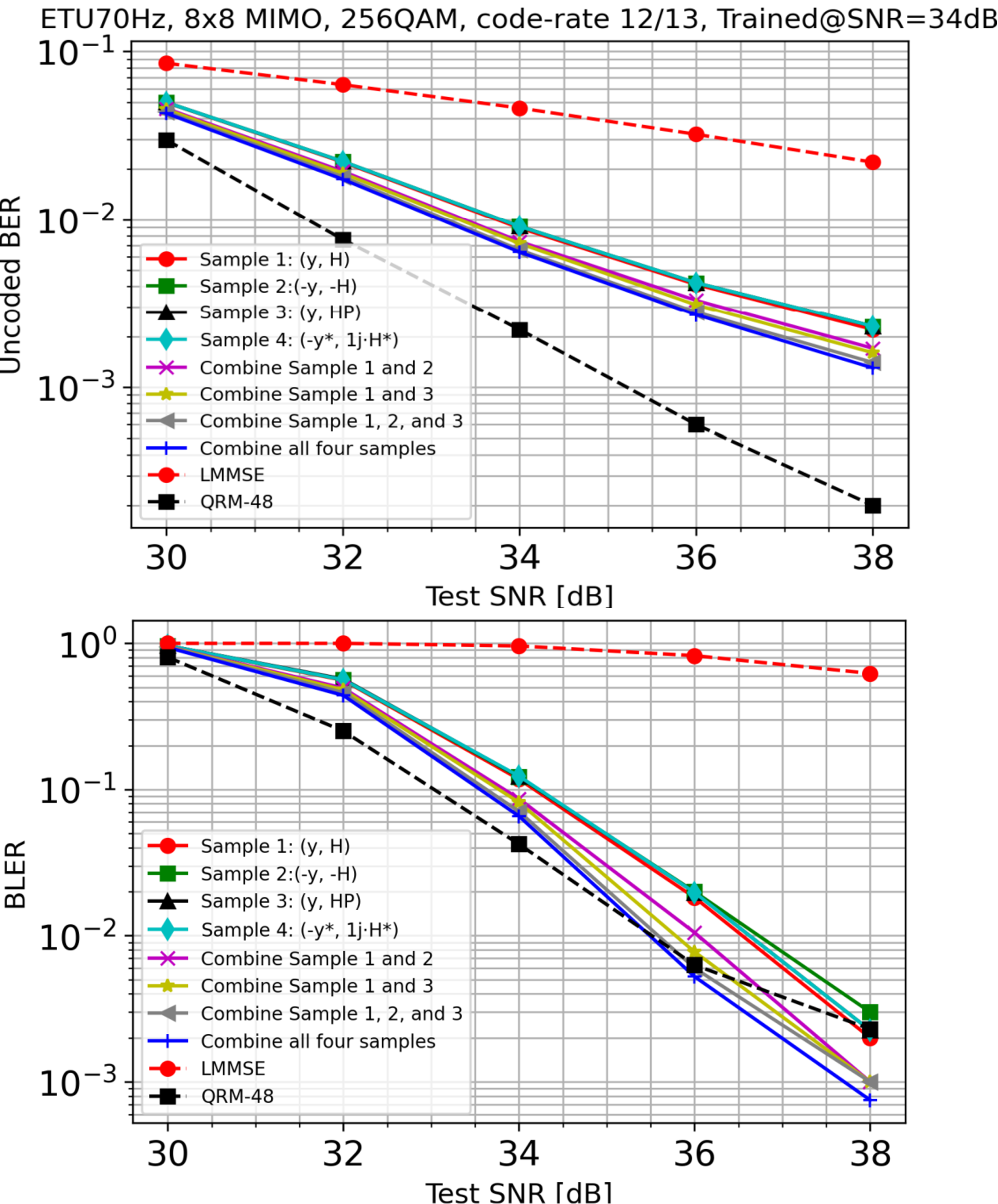}
        \vspace{-3mm}
    \caption{Uncoded BER and BLER for $8\!\times\! 8$ MIMO and 256QAM modulation under ETU-70Hz channel. }  
    \label{fig:AIRec}
    \vspace{-6mm}
\end{figure} 


\subsection{Performance Enhancements}
Next we consider an $8\!\times\! 8$ MIMO with 256QAM modulation in (\ref{mm}). In addition, a low-density parity-check (LDPC) code with a high code-rate $12/13$ is applied for measuring the block-error-rate (BLER). The channel $\vec{H}$ is modeled as ETU-70Hz specified in 5G. The AI model is trained at SNR=34dB, and in this case, directly output LLR values $L_{m,k}$ for decoder. We resample the AI model with three invariant transformations: (-\vec{y}, -\vec{H}), (\vec{y}, \vec{H}\vec{P}), and $(\vec{y}^{\ast}, 1j\!\cdot\!\vec{H}^{\ast)}$, where $\vec{P}$ denotes a random column permutation on $\vec{H}$. 

As can be seen in Fig.~5, although the AI model is trained at SNR=34dB, it works well for SNR values from 30dB to 38dB. By combing the inference outputs from multiple samples, both the uncoded BER and BLER are consistently improved. With four inference outputs, there is around 0.5dB gains at 1\% uncoded BER, and 0.7dB gains at 10\% BLER. These gains become larger as SNR increases. In this case, the ML detector is infeasible and we compare AI detector to  conventional LMMSE and QRM detectors. While LMMSE has the worst performance, the AI detector approaches the QRM detector ~\cite{Im2007QRMMLD} that preserves 48 best nodes on each layer after sorting $48\!\times\!256$ candidate nodes. As shown in Fig.~3 and also from Theorem~1, the gains with combining more samples become smaller as $M$ increases.

\section{Summary}  

We have proposed a resampling technique for a trained AI model. Under invariant transformations, the trained AI model produces inference outputs with errors that may have independencies among transformed input samples . Combing multiple inference outputs can effectively improve the inference accuracy and reduce the epistemic uncertainty. We have demonstrated the proposal with an AI based MIMO detector design, and showed that under $8\!\times\! 8$ MIMO and 256QAM modulation, the SNR gains with resampling the trained AI model can be up to 1dB.

\newpage

\bibliographystyle{IEEEtran}

\end{document}